\documentclass[]{TEAI}
\usepackage{helvet}

\usepackage{amsmath} 
\usepackage{natbib}
\usepackage{graphicx}
\usepackage{subcaption} 
\usepackage[table,dvipsnames]{xcolor}         

\usepackage[utf8]{inputenc} 
\usepackage[T1]{fontenc}    
\definecolor{fudanblue}{rgb}{0.055,0.254,0.610}
\definecolor{kleinblue}{rgb}{0,0.22,0.65}
\definecolor{green}{rgb}{0.0,0.265,0.0}

\usepackage{url}            
\usepackage{booktabs}       
\usepackage{amsfonts}       
\usepackage{nicefrac}       
\usepackage{microtype}      

\usepackage{amssymb}
\usepackage{enumitem}
\usepackage{amsmath}
\usepackage{xspace}
\usepackage{multirow}
\usepackage{bbding}
\usepackage{booktabs}       

\usepackage{array}
\usepackage{relsize}

\newcommand{\ourrepa}{REPA-A\xspace}
\newcommand{\system}{\textsc{CaTok}\xspace}
\usepackage{arydshln}

\crefname{section}{Sec.}{Secs.}
\Crefname{section}{Section}{Sections}
\Crefname{table}{Table}{Tables}
\crefname{table}{Tab.}{Tabs.}
\Crefname{figure}{Figure}{Figures}
\crefname{figure}{Fig.}{Figs.}
\Crefname{equation}{Equation}{Equations}
\crefname{equation}{Eq.}{Eqs.}
\usepackage{cuted}
\usepackage{capt-of}

\newcommand{\tablestyle}[2]{\setlength{\tabcolsep}{#1}\renewcommand{\arraystretch}{#2}\centering}

\usepackage{makecell}

\def\eg{\emph{e.g.}\xspace} 

\def\ie{\emph{i.e.}\xspace} 
 
\def\vs{\emph{vs.}\xspace}

\usepackage[toc,page,header]{appendix}
\usepackage[utf8]{inputenc} 
\usepackage[T1]{fontenc}    
\usepackage{hyperref}       
\usepackage{url}            
\usepackage{booktabs}       
\usepackage{lmodern}        
\usepackage{amsfonts}       
\usepackage{nicefrac}       
\usepackage{microtype}      
\usepackage{wrapfig}

\usepackage{amssymb}  
\usepackage{fontawesome}  
\usepackage{url}  

\usepackage{titletoc}

\usepackage{tikz}  
\usepackage{comment}  
\usepackage{tabularx}  
\usepackage{booktabs}  

\usepackage{minitoc}

\usepackage{booktabs}
\usepackage{array}
\usepackage{etoolbox}

\definecolor{lightblue}{RGB}{200, 230, 255}  
\definecolor{headerblue}{RGB}{150, 200, 255} 

\usepackage{pgfplots}
\usepackage[utf8]{inputenc} 
\usepackage[T1]{fontenc}    
\usepackage{hyperref}       
\usepackage{url}            
\usepackage{booktabs}       
\usepackage{amsfonts}       
\usepackage{nicefrac}       
\usepackage{microtype}      
\usepackage{xcolor}         
\usepackage{graphicx}
\usepackage{float}
\usepackage{comment}
\usepackage{multirow} 
\usepackage{amsmath} 
\usepackage{makecell} 
\usepackage{siunitx}  
\usepackage{tikz}
\usepackage{pgf-pie} 
\usepackage{subcaption}
\usepackage{wrapfig}
\usepackage[export]{adjustbox}

\usepackage{ragged2e}      
\usepackage{tabularx}       
\usepackage{array}          
\usepackage{caption}        
\usepackage{enumitem}
\usepackage{pifont}
\usepackage[hang,flushmargin]{footmisc} 

\usepackage{tcolorbox}

\usepackage{tcolorbox}    
\tcbuselibrary{breakable}  
\tcbuselibrary{skins}      

\usepackage{tabularx}
\usepackage{listings}

\title{\system: Taming Mean Flows\\for One-Dimensional Causal Image Tokenization}

\author{
    Yitong Chen\textsuperscript{\rm1,\rm2,\rm3}\quad 
    Zuxuan Wu\textsuperscript{\rm1,\rm2,\rm3,$\dagger$}\quad 
    Xipeng Qiu\textsuperscript{\rm1,\rm2,\rm3}\quad 
    Yu-Gang Jiang\textsuperscript{\rm1,\rm3,$\dagger$}
}
\affiliation[1]{\mbox{ Institute of Trustworthy Embodied AI, Fudan University}} 
\affiliation[2]{\mbox{ Shanghai Innovation Institute}}
\affiliation[3]{\mbox{ Shanghai Key Laboratory of Multimodal Embodied AI}}

\teaser{%
  \includegraphics[width=\textwidth]{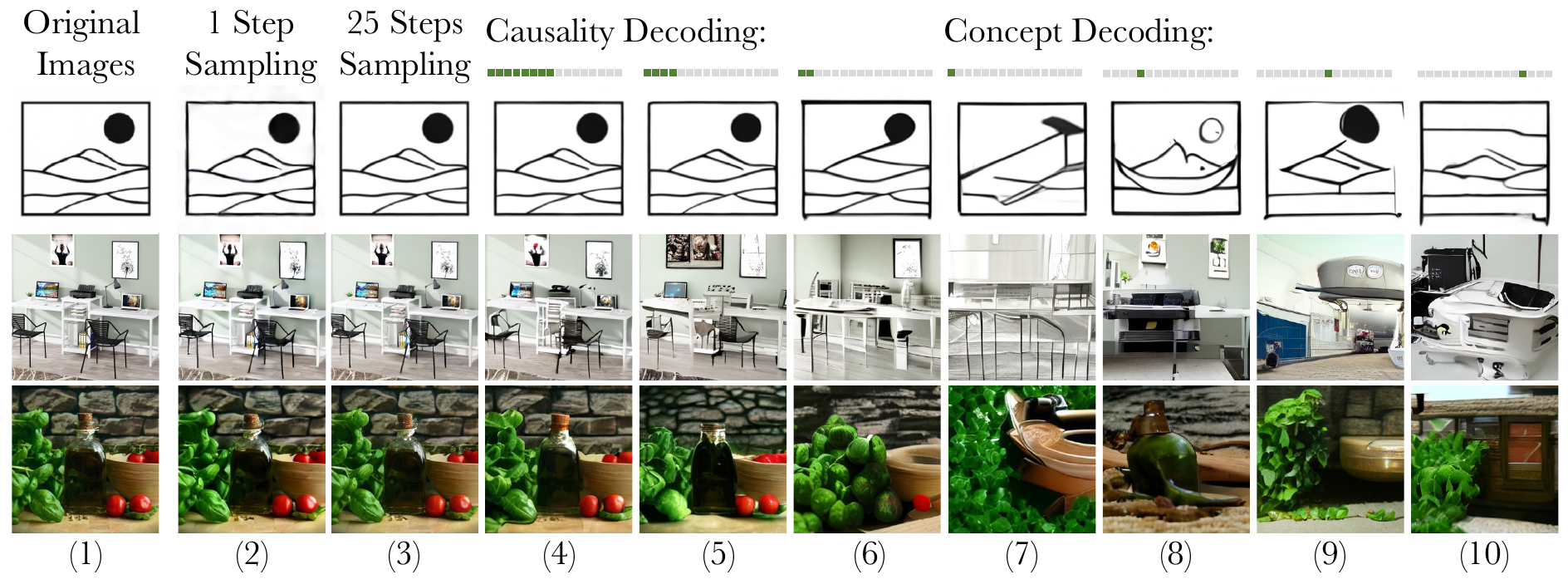}
  \captionof{figure}{
    \textbf{Reconstruction samples.} \system with a MeanFlow decoder~\citep{meanflow} supports fast one-step (col. 2) and high-quality multi-step (col. 3) sampling with 256 tokens. Reconstructions in cols. 3–7 show a fine-to-coarse trend as tokens are reduced from 256 to 16, highlighting the causality of the 1D tokens. Cols. 7–10 present reconstructions from different 16-token segments, demonstrating that \system naturally learns diverse visual concepts across token intervals.\\ \vspace{-3.3mm}
  }
  \label{fig:teaser}
}

\abstract{
\begin{abstract}
AAutoregressive (AR) language models rely on causal tokenization, but extending this paradigm to vision remains non-trivial. Current visual tokenizers either flatten 2D patches into non-causal sequences or enforce heuristic orderings that misalign with the ``next-token prediction'' pattern. Recent diffusion autoencoders similarly fall short: conditioning the decoder on all tokens lacks causality, while applying nested dropout mechanism introduces imbalance. To address these challenges, we present \system, a 1D causal image tokenizer with a MeanFlow decoder. By selecting tokens over time intervals and binding them to the MeanFlow objective, as illustrated in \cref{fig:teaser}, \system learns causal 1D representations that support both fast one-step generation and high-fidelity multi-step sampling, while naturally capturing diverse visual concepts across token intervals. To further stabilize and accelerate training, we propose a straightforward regularization REPA-A, which aligns encoder features with Vision Foundation Models (VFMs). Experiments demonstrate that \system achieves state-of-the-art results on ImageNet reconstruction, reaching 0.75 FID, 22.53 PSNR and 0.674 SSIM with fewer training epochs, and the AR model attains performance comparable to leading approaches. Project website is available in \url{https://sharelab-sii.github.io/catok-web}.
\end{abstract}
}

\begin{document}
\maketitle

\renewcommand{\thefootnote}{}
\footnotetext{$^\dagger$Corresponding authors.}


\vspace{-1.5em}

\begin{figure*}[ht!]
    \centering
    \includegraphics[width=0.8  \linewidth]{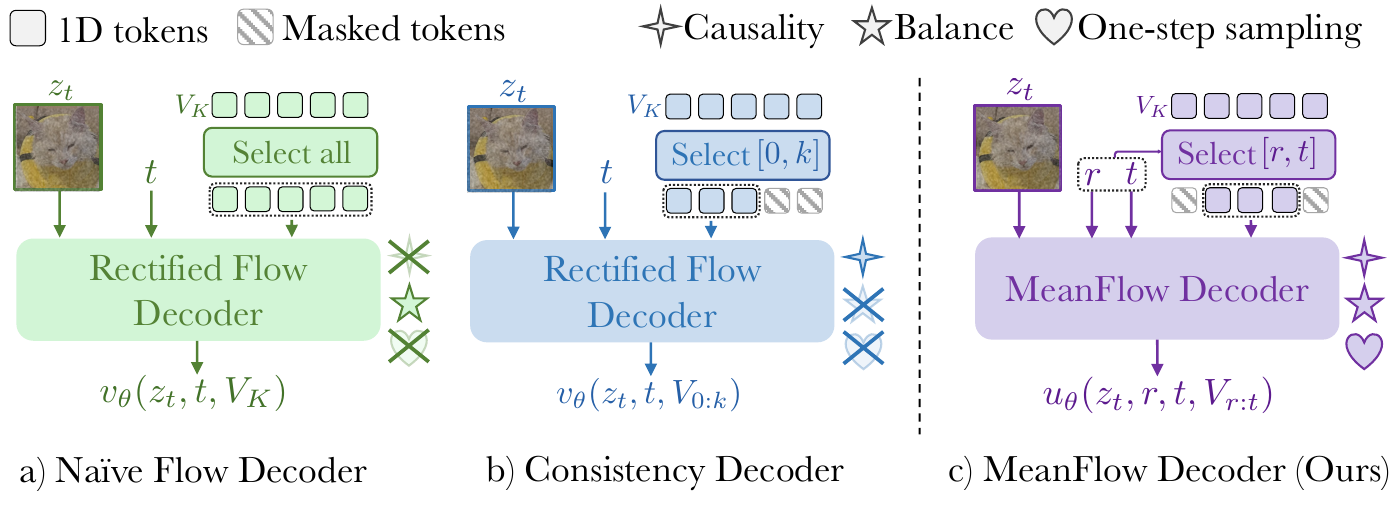}
    \caption{
    \textbf{Comparison among different decoders.} \textbf{a)} Na\"ive flow decoders~\citep{flowmo} condition on all 1D tokens from the encoder without dropout, leading the 1D tokens to lack causality; \textbf{b)} Consistency decoders obtain $k$ by random sampling~\citep{flextok,semanticist} or timestep binding~\citep{ddt,selftok}, and condition on the first $k$ 1D tokens, which biases toward early tokens, introducing imbalance, leading to degraded performance of AR generation; \textbf{c)} Our MeanFlow decoder conditions on 1D tokens within the time interval $[r,t]$ to model the average velocity field along the subpath, which inherently maintains \textbf{causality} and \textbf{balance} of the 1D visual tokens, and supporting \textbf{one-step sampling} during image reconstruction or generation.}
    \label{fig:compare}
\end{figure*}

\section{Introduction}
The autoregressive (AR) paradigm enables generative large language models (LLMs) to achieve remarkable progress, exhibiting strong generalization and scalability~\citep{gpt4,gpt5,gemini,llama3,qwen3,deepseek}. Following the natural reading order of the text, LLMs tokenize a sentence into 1D causal tokens and perform generative modeling through next-token prediction. To emulate the capabilities and properties of LLMs in visual generation, the computer vision community has recently advanced large autoregressive vision models~\citep{lvm,magvitv2,emu3,llamagen,infinity,simplear}. However, due to inferior performance, diffusion-based models~\citep{ddpm,score-based} like rectified flows~\citep{rectified-flow,flowmatching} remain the dominant approach in most scenarios~\citep{sora,qwen-image}.

In this paper, we argue that a crucial step toward bridging the gap between autoregressive language models and vision models lies in the causal tokenization of visual content. Autoregressive modeling relies on causal tokens and requires a predefined order of data. Unlike text, which inherently possesses a natural order, defining an appropriate order for images remains an open issue. VQGAN-like models~\citep{vqgan,dualtoken,omnitokenizer} tokenize an image into grids of 2D tokens, and flatten them to a 1D sequence in raster~\citep{vqvae-2,dalle} or random~\citep{rar,mar} order, which lacks causality between preceding and succeeding tokens~\citep{ddt,selftok}. VAR-like models~\citep{var,infinity}, on the other hand, tokenize images into multi-scale 2D tokens and establish a coarse-to-fine ordering via next-scale prediction. While this approach guarantees causality in visual tokens and yields promising results, it compromises the ``next-token prediction'' pattern of LLMs.

With the recent advances in 1D tokenizers~\citep{seed,titok}, the community has renewed its interest in diffusion autoencoders~\citep{diff-ae, cdiff-ae, ssdd} due to their demonstrated effectiveness in visual generation. Diffusion autoencoders extract 1D tokens from registers~\citep{registers} of encoders, and use them as conditions for the decoder to reconstruct images with denoising or rectified flow objective. However, as shown in \cref{fig:compare} a), Na\"ive flow decoders, such as FlowMo~\citep{flowmo}, condition on all 1D tokens from the encoder, causing the 1D tokens to lack causality and making AR learning difficult. To learn the causality for 1D tokens, as shown in \cref{fig:compare} b), consistency decoders apply nested dropout~\citep{nestdrop} by conditioning on the first $k$ tokens, where $k$ is determined either via random sampling, as in FlexTok~\citep{flextok} and Semanticist~\citep{semanticist}, or via timestep binding, as in DDT-Llama~\citep{ddt} and Selftok~\citep{selftok}. Since earlier tokens are more likely to be selected, this approach introduces imbalance and can be harmful to AR generation (see \cref{tab:abl_causal_balance}).

Motivated by these observations, we propose \system, a 1D \textbf{\textsc{Ca}}usal image \textbf{\textsc{Tok}}enizer equipped with a MeanFlow decoder~\citep{meanflow}. As illustrated in \cref{fig:compare} c), we address the imbalance problem by selecting 1D tokens within a sampled time interval $[r,t]$ and binding them with the corresponding time interval in the MeanFlow objective. This allows the 1D tokens to model the average velocity field along the subpath from $r$ to $t$, capturing causality in the noise-to-image generation process while naturally supporting one-step sampling during generation. Moreover, inspired by REPA and REPA-E~\citep{repa, repa-e}, we align the image features from encoders with high-quality external visual representations, providing a regularization that effectively accelerates and stabilizes autoencoder training. We refer to this variant as \ourrepa.

As shown in \cref{fig:teaser}, \system supports both fast one-step sampling (col. 2) and high-quality multi-step sampling (col. 3) with 256 tokens, demonstrating its flexibility in balancing efficiency and fidelity. Reconstructions in cols. 3–7, obtained by progressively reducing the number of tokens from 256 to 16, exhibit a clear fine-to-coarse trend, providing evidence for the causality of the learned 1D tokens. Moreover, reconstructions in cols. 7–10 from different 16-token segments show that \system naturally learns diverse visual concepts across token intervals, underscoring its ability to disentangle semantic information and distribute it meaningfully among tokens. Our contributions can be summarized as:

\begin{enumerate}[leftmargin=2em]
  \item
  We propose a novel architecture for 1D causal image tokenization based on diffusion autoencoders~\citep{diff-ae} with the MeanFlow~\citep{meanflow} objective.
    \item 
    We seamlessly combine the training of a causal encoder and a one-step flow decoder, enabling one-step sampling in diffusion autoencoders.
    \item 
    We propose \ourrepa, an advanced technique that leverages existing vision foundation models to stabilize and accelerate diffusion autoencoder training.
    \item 
    On ImageNet, our \system-L achieves state-of-the-art results with 0.75 rFID, 22.53 PSNR and 0.674 SSIM, while attains comparable performance to leading approaches with 2.95 gFID.
\end{enumerate}
\section{Background}
In this section, we provide a concise introduction to rectified flows~\citep{rectified-flow, flowmatching} and MeanFlow models~\citep{meanflow}.
\subsection{Rectified flows}
Given data $x \sim p_{data}(x)$ and prior $\epsilon \sim p_{prior}(\epsilon)$, rectified flows learn the conditional velocity fields $v_t=v_t(z_t|x)$ between these two distributions. Specifically, a flow path can be constructed as $z_t = (1-t)x + t\epsilon$ with time $t$, and the conditional velocity can be derived by:
\begin{equation}
    v(z_t| x) =  \frac{d}{dt}z_t = \epsilon - x.
    \label{eq:cfm}
\end{equation}
A deep neural network $v_\theta(z_t, t)$ parameterized by $\theta$ is learned to model the marginal velocity field
\begin{equation}
    v(z_t, t) \triangleq \mathbb{E}_{p_t(v_t|z_t)}[v_t],
\end{equation}
which is equivalent to fitting the conditional velocity field in \cref{eq:cfm}~\citep{flowmatching}. In inference, starting from $z_1 = \epsilon \sim p_{prior}(\epsilon)$, samples can be generated by solving:
\begin{equation}
    z_r = z_t - \int_r^tv_{\theta}(z_{\tau}, \tau)d\tau,
\end{equation}
where $r$ denotes another timestep and $r<t$. In practice, this integral is numerically approximated in discrete time steps. For instance, the Euler method updates each step as:
\begin{equation}
    z_r = z_t - (t-r)v_{\theta}(z_t, t).
    \label{eq:el1}
\end{equation}
However, it estimates the average velocity over the interval $[r, t]$ using only the instantaneous velocity at time $t$, which introduces inaccuracies during sampling. 

\subsection{MeanFlow models}
To mitigate the errors that arise with fewer sampling steps, MeanFlow models directly fit the average velocity $u$ over the interval $[r, t]$. Formally, the average velocity $u$ can be defined as:
\begin{equation}
    u(z_t, r, t) \triangleq \frac{1}{t-r} \int_r^tv(z_{\tau}, \tau)d\tau.
    \label{eq:u}
\end{equation}
Through derivations in \cite{meanflow}, the average velocity $u(z_t, r, t)$ can be obtained from the instantaneous velocity:
\begin{equation}
    u(z_t, r, t) = v(z_t, t) - (t-r)(v(z_t, t)\partial_zu(z_t, r, t)+\partial_tu(z_t, r, t)),
\end{equation}
and the MeanFlow objective is:
\begin{equation}
    \mathcal{L}(\theta) = \mathbb{E}||u_\theta - \text{sg}[v(z_t|x)-(t-r)(v(z_t|x)\partial_zu_\theta+\partial_tu_\theta)]||_2^2,
    \label{eq:mf_obj}
\end{equation}

where $\text{sg}[\cdot]$ denotes the stop-gradient operation, avoiding double backpropagation through the Jacobian–vector product. Moreover, one-step sampling can be given by $z_0 = \epsilon - u_\theta(\epsilon, 0, 1)$.

\begin{figure*}[t!]
    \centering
    \includegraphics[width=0.8\linewidth]{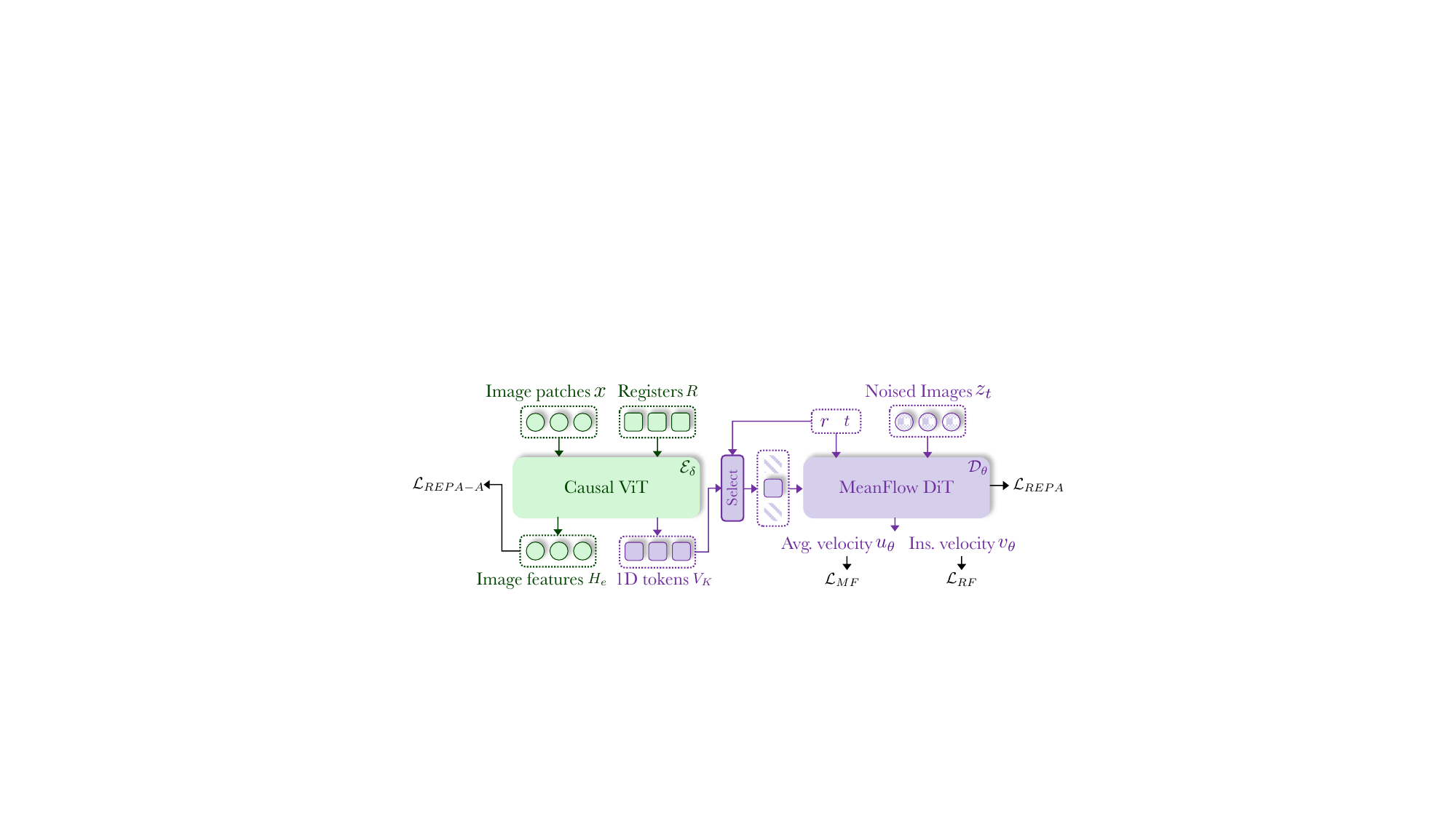}
    \caption{\textbf{Architecture of our \system.} \system is a diffusion autoencoder with a causal Vision Transformer (ViT)~\citep{vit} encoder and a MeanFlow Diffusion Transformer (DiT)~\citep{dit} decoder. The encoder leverages registers~\citep{registers} to extract rich visual information into 1D tokens, which are then conditioned to the decoder through time interval selecting. With two flow objectives and two representation alignment objectives, \system learns causal 1D representations that support both one-step and multi-step sampling, while naturally capturing diverse visual concepts across different token intervals.}
    \label{fig:arch}
\end{figure*}

\section{\system}
We now introduce \system, a diffusion autoencoder~\citep{diff-ae,cdiff-ae} with a causal Vision Transformer (ViT)~\citep{vit} encoder and a MeanFlow Diffusion Transformer (DiT)~\citep{dit} decoder, for 1D causal image tokenization. We begin in \cref{sec:arch} by presenting the architecture of \system. Next, in \cref{sec:train}, we describe how it is optimized through multiple objectives. Finally, in \cref{sec:infer}, we outline the autoregressive modeling used for image generation with the trained \system.

\subsection{Architecture}
\label{sec:arch}
As shown in \cref{fig:arch}, \system is a diffusion autoencoder with a causal ViT encoder $\mathcal{E}_\delta$ and a MeanFlow DiT decoder $\mathcal{D}_\theta$ parameterized by $\delta$ and $\theta$ respectively. Specifically, given an image $x$, we concatenate it with $K$ registers $R$ and send them into the encoder:
\begin{equation}
    [H_e, V_K] = \mathcal{E}_\delta([x, R]),
\end{equation}
where $H_e$ denotes the image features and $V_K$ represents the compressed 1D tokens. Furthermore, a causal attention mask is applied to enforce the dependency structure among 1D tokens~\citep{seed,flextok,semanticist}. Specifically, image features can attend to each other but not to the 1D tokens; in contrast, 1D tokens are allowed to attend to all image features while being restricted to only their preceding 1D tokens.

In the MeanFlow DiT decoder phase, we first independently sample two timesteps $r$ and $t$, ensuring that $r,t \in [0,1]$ and $r < t$. Then, the flow path is constructed by linearly interpolating the image $x$ with random noise $\epsilon \sim \mathcal{N}(0,1)$:
\begin{equation}
    z_t = (1-t)x + t\epsilon.
\end{equation}
By conditioning the noised image $z_t$ with the 1D tokens from the interval $[r \cdot K, t \cdot K]$, denoted as $V_{r:t}$, and timesteps $r, t$, the DiT decoder predicts the average velocity $u_\theta$ over the time interval:
\begin{equation}
    u_\theta = \mathcal{D}_\theta(z_t, r, t, V_{r:t}).
    \label{eq:u_avg}
\end{equation}
Since accurately modeling the instantaneous velocity field improves training stability when learning the average velocity field~\citep{meanflow,facm}, we follow \cref{eq:u} and set $r = t$ to model the instantaneous velocity field $v_\theta$:
\begin{equation}
    v_\theta = \mathcal{D}_\theta(z_t, t, t, V_K),
\end{equation}
and all the 1D tokens $V_K$ are conditioned upon.

\subsection{Training}
\label{sec:train}
As illustrated in \cref{fig:arch}, \system is jointly optimized with two flow objectives—MeanFlow~\citep{meanflow} and Rectified Flow~\citep{rectified-flow,flowmatching}—and two representation alignment objectives—REPA~\citep{repa} and our proposed \ourrepa.

\textbf{MeanFlow objective.} From \cref{eq:cfm}, \cref{eq:mf_obj} and \cref{eq:u_avg}, the MeanFlow objective is defined as:
\begin{equation}
    \mathcal{L}_{MF} := \mathbb{E}||u_\theta - (\epsilon - x)- \text{sg}[(t-r)((\epsilon - x)\partial_zu_\theta+\partial_tu_\theta)]||_2^2,
    \label{eq:loss_mf}
\end{equation}
where $\text{sg}[\cdot]$ denotes the stop-gradient operation, and $(\epsilon - x)\partial_zu_\theta+\partial_tu_\theta$ is computed using the Jacobian-vector product operation.

\textbf{Rectified Flow objective.} We also model the instantaneous velocity field to enhance training stability. Based on \cref{eq:cfm}, we define our Rectified Flow objective as follows:
\begin{equation}
    \mathcal{L}_{RF} := \mathbb{E}||v_\theta - (\epsilon - x)||_2^2.
    \label{eq:loss_rf}
\end{equation}
Following \cite{meanflow}, we employ an adaptive $L_2$ loss in place of the standard $L_2$ loss to enhance performance, defined as $\mathcal{L}_{\text{adaptive}} = ||\Delta||_2^2\ /\ \text{sg}[(||\Delta||_2^2+c)^w]$, where $\Delta$ denotes the regression error, and integrate the two objectives in $\mathcal{L}_{F}$ by fixing a proportion $q$ of samples with $r = t$. In our implementation, we set $c = 10^{-3}$, $w = 1.0$, and $q = 75\%$.

\textbf{REPA objective.} REPA~\citep{repa} is a regularization technique that leverages Vision Foundation Models (VFMs) to assist DiT training and accelerate convergence. Formally, given the hidden states $H_d$ from a middle layer of the DiT decoder and pretrained representations $H_{vfm}$ from a VFM, our REPA objective can be defined as:
\begin{equation}
    \mathcal{L}_{REPA} := -\mathbb{E}[\frac{1}{N}\sum_{n=1}^N\text{sim}(H_{vfm}^{[n]},\text{proj}(H_d^{[n]}))],
    \label{eq:loss_repa}
\end{equation}
where $n$ is a patch index, $\text{sim}(\cdot,\cdot)$ is the cosine similarity function and $\text{proj}(\cdot)$ is the projection layer.

\textbf{Our proposed \ourrepa objective.} Unlike REPA-E~\citep{repa-e}, which backpropagates gradients to the VAE~\citep{vae}, or VA-VAE~\citep{va-vae}, which directly regularizes the compressed features of VAE using VFMs, we propose \ourrepa, a representation alignment method specifically tailored for conditional diffusion autoencoders such as our \system. Formally, given the image features $H_e$ from the encoder and the VFM representations $H_{vfm}$, the \ourrepa can be defined as:
\begin{equation}
    \mathcal{L}_{REPA-A} := -\mathbb{E}[\frac{1}{N}\sum_{n=1}^N\text{sim}(H_{vfm}^{[n]},H_e^{[n]})],
    \label{eq:loss_repa_a}
\end{equation}
where $n$ is a patch index and $\text{sim}(\cdot,\cdot)$ is the cosine similarity function. With \ourrepa, the encoder produces higher quality semantic representations, allowing 1D tokens to extract more informative and discriminative visual content, thereby accelerating convergence and enhancing overall performance.

\label{apend:ar}
\begin{figure*}[t!]
    \centering
    \includegraphics[width=0.8\linewidth]{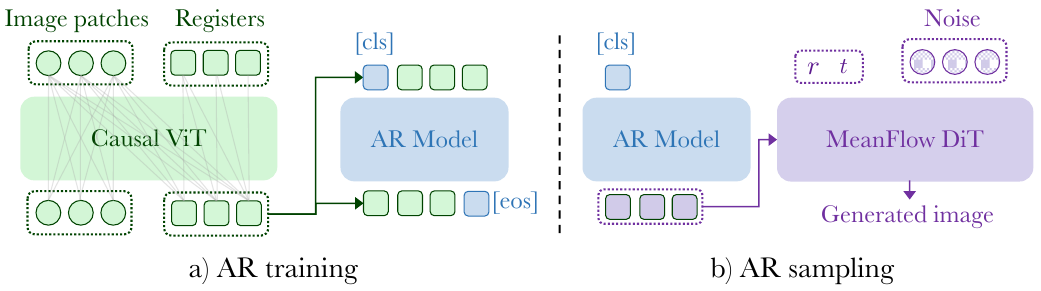}
    \caption{We visualize the causal mask mechanism in ViT in a). After training \system, we freeze the encoder to extract 1D tokens. During AR training stage, these tokens are optimized with a class token prefix using teacher forcing under a diffusion loss~\citep{mar}. At sampling time, we input a learned class token, the AR model predicts the corresponding visual 1D tokens, and these tokens are then conditioned to the decoder for generation.}
    \label{fig:arch_ar}
\end{figure*}

\subsection{Autoregressive modeling}
As shown in \cref{fig:arch_ar}, once the causal 1D tokens $V_K$ are obtained from a well-trained \system encoder, we train a standard autoregressive model following the “next-token prediction” paradigm to generate images. Formally, the AR model defines the generation process as:
\begin{equation}
    p(V_1, V_2,...,V_k) = \prod \limits_{k=1}^Kp(V_k | V_1,...,V_k).
    \label{eq:ar}
\end{equation}
When $V_k$ is represented as discrete indices, this probabilistic model can be optimized via cross-entropy loss. In contrast, when $V_k$ is continuous-valued, as in our setting, optimization is performed using a diffusion loss introduced in \cite{mar}.

\label{sec:infer}
For image generation, given a prior such as a class token, we first obtain a predicted sequence ${\hat{V}_K}$ via \cref{eq:ar}. By feeding it into MeanFlow decoder, we can directly perform one-step sampling to render an image through $\hat{x} = \epsilon - \mathcal{D}_\theta(\epsilon, 0, 1, \hat{V}_K)$, where $\epsilon$ denotes a random Gaussian noise.

\section{Related works}
AR modeling requires compressing raw data into a sequence of tokens, which in turn has spurred a line of research on visual tokenizers. We categorize them into three types.

\textbf{2D visual tokenizers.} VQ-VAE~\citep{vq-vae,vqvae-2} is one of the most widely adopted 2D visual tokenizers, integrating Vector Quantization (VQ) into the VAE~\citep{vae} to produce discrete tokens from image patches. Subsequent works improve upon this design: VQGAN~\citep{vqgan} introduces an adversarial loss to enhance reconstruction quality, while RQ-VAE~\citep{rq-vae} employs multiple quantization stages. MAGVIT-v2~\cite{magvitv2} further alleviates quantization bottlenecks with Look-up Free Quantization (LFQ), and MaskBit~\cite{maskbit} modernizes the VQGAN framework with binary quantized tokens. Most recently, VAR models~\citep{var,infinity} tokenize images into multi-scale 2D tokens and establish a coarse-to-fine ordering via ``next-scale prediction''. However, these 2D tokenizers either lack causality or compromise the ``next-token prediction'' paradigm.

\textbf{1D visual tokenizers.} SEED~\citep{seed} employs a causal Q-Former~\citep{blip-2} to extract 1D tokens from a ViT~\cite{vit} encoder and performs semantic reconstruction with a pre-trained text encoder. TiTok~\citep{titok} derives 1D tokens using learnable registers and conditions a ViT decoder for mask-to-patch reconstruction. Building on these designs, a line of work explores 1D causal visual tokenizers. TexTok~\citep{textok} and TA-TiTok~\citep{ta-titok} leverage textual conditioning to enhance performance, ALIT~\citep{alit} introduces adaptive-length tokenization via recurrent encoding, One-D-Piece~\citep{oned-piece} applies nested dropout~\citep{nestdrop} on tokens to introduce causality, and SpectralAR~\citep{spectralar} adopts a similar architecture but imposes explicit spectral interpretations to supervise different tokens. In contrast, \system adopts a diffusion-based decoder, which we introduce next.

\textbf{Diffusion autoencoders as 1D tokenizers.} Diffusion autoencoders~\citep{diff-ae,cdiff-ae,rcg,vilex} compress image features into 1D tokens, which serve as conditioning inputs for diffusion models trained with denoising or rectified flow objectives. However, naïve flow decoders such as FlowMo~\citep{flowmo} and DiTo~\citep{dito} condition on all tokens simultaneously, eliminating causal structure and thereby hindering AR learning. To address this, consistency decoders introduce causality through nested dropout, conditioning only on early tokens. The early-token set is determined either stochastically, as in FlexTok~\citep{flextok} and Semanticist~\citep{semanticist}, or deterministically via timestep binding, as in DDT~\citep{ddt} and Selftok~\citep{selftok}. However, because earlier tokens are disproportionately favored, these methods induce imbalance, which can degrade AR generation quality. In contrast, our \system leverages an additional MeanFlow~\citep{meanflow} objective to capture visual causality in a balanced manner while naturally supporting one-step sampling.
\begin{table*}[!ht]
      \centering
      \tablestyle{8pt}{0.9}
      \caption{ \textbf{Reconstruction results on ImageNet 256$\times$256 benchmark.} ``Token'' denotes the number of tokens used for reconstruction, and ``Dim.'' denotes the dimension of these tokens. ``Param.'' indicates the model size, and ``VQ'' specifies whether the tokens are vector-quantized. ``$\downarrow$'' or ``$\uparrow$'' denote lower or higher values are better. $\dagger$: enabling one-step sampling. $\star$: trained by official codebase~\cite{semanticist}.}
      
      \begin{tabular}{lcccccccc}
      \toprule
      Method & Token & \#Dim. & \#Param.  & Epochs & VQ & rFID$\downarrow$ & PSNR$\uparrow$ & SSIM$\uparrow$  \\
      \midrule
      \multicolumn{9}{c}{\textcolor{gray}{\textit{One-step 2D tokenizers}}} \\ 
      VQGAN~\cite{vqgan} & 16x16 & 16 & 307M & - & \checkmark & 7.94 & - & - \\
      LlamaGen~\cite{llamagen} & 16x16 & 8 &72M & 40 & \checkmark & 2.19 & 20.67 & 0.589 \\
      MaskBit~\cite{maskbit} & 16x16 & 12 &54M & 270 & \checkmark & 1.37 & 21.50 & 0.560 \\
      MAR-VAE~\cite{mar} & 16x16 & 16 & - & - & $\times$ & 1.22 & - & - \\
      OpenMagViT-V2~\cite{openmagvit2} & 16x16 & - & 116M & 270 & \checkmark & 1.17 & 21.63 & 0.640 \\
      \midrule
      \multicolumn{9}{c}{\textcolor{gray}{\textit{One-step 1D tokenizers}}} \\ 
      SpectralAR-64~\cite{spectralar} & 64 & 16 &  172M & 300 & \checkmark & 4.03 & - & - \\
      TiTok-S-128~\cite{titok} & 128 & 16 & 44M & 300 & \checkmark & \uline{1.71} & 17.52 & 0.437 \\
      TiTok-L-32~\cite{titok} & 32 & 8 & 614M & 300 & \checkmark & 2.21 & 15.60 & 0.359 \\
      One-D-Piece-B-256~\cite{oned-piece} & 256 & 16 & 172M & 300 & \checkmark & \textbf{1.11} & 18.77 & - \\
      \rowcolor{OliveGreen!5}
      \system-B-256$^\dagger$ & 256 & 16 & 224M & 80 & $\times$ & 4.89 & \uline{20.77} & \uline{0.617} \\
      \rowcolor{OliveGreen!5}
      \system-L-32$^\dagger$ & 32 & 16 & 552M & 160 & $\times$ & 4.48  & 17.25 & 0.441 \\
          \rowcolor{OliveGreen!5}
      \system-L-64 $^\dagger$ & 64 & 16 & 552M & 160 & $\times$ &  4.96 & 18.56 & 0.506 \\
      \rowcolor{OliveGreen!5}
      \system-L-128$^\dagger$ & 128 & 16 & 552M & 160 & $\times$ & 4.92 & 20.36 &  0.590 \\
      \rowcolor{OliveGreen!5}
      \system-L-256$^\dagger$ & 256 & 16 & 552M & 160 & $\times$ & 4.63 &  \textbf{20.99} &  \textbf{0.630}\\
      \midrule
      \multicolumn{9}{c}{\textcolor{gray}{\textit{Diffusion tokenizers}}} \\ 
      FlexTok d12-d12~\cite{flextok} & 256 & 6 & 170M & 640 & \checkmark & 4.20 & - & - \\
      FlexTok d18-d18~\cite{flextok} & 256 & 6 & 573M & 640 & \checkmark & 1.61 & - & - \\
      FlexTok d18-d28~\cite{flextok} & 256 & 6 & 1.4B & 640 & \checkmark & 1.45 & 18.53 & 0.465 \\
      Semanticist-L-256~\cite{semanticist} & 256 & 16 & 552M   & 400 & $\times$ & \uline{0.78} & 21.61 & 0.626 \\
      Semanticist-L-128$^\star$~\cite{semanticist} & 256 & 16 & 552M   & 400 & $\times$ & 1.24 & 19.59 & 0.586 \\
      SelfTok-512~\cite{selftok} & 512 & 16 & - & - & \checkmark & - & 21.86 & 0.600 \\
      FlowMo-Lo-256~\cite{flowmo} & 256 & - & 945M & 130 & \checkmark & 0.95 & 22.07 & 0.649 \\
      \rowcolor{OliveGreen!5}
      \system-B-256 & 256 & 16 & 224M & 80 & $\times$ & 1.17 & \uline{22.10} & \uline{0.666} \\
      \rowcolor{OliveGreen!5}
      \system-L-32 & 32 & 16 & 552M & 160 & $\times$ & 2.03 &  17.85 & 0.465 \\
      \rowcolor{OliveGreen!5}
      \system-L-64 & 64 & 16 & 552M & 160 & $\times$ & 1.61 &  19.36 & 0.533 \\
      \rowcolor{OliveGreen!5}
      \system-L-128 & 128 & 16 & 552M & 160 & $\times$  &  1.17 & 21.01  & 0.609  \\
      \rowcolor{OliveGreen!5}
      \system-L-256 & 256 & 16 & 552M & 160 & $\times$ & \textbf{0.75} & \textbf{22.53} & \textbf{0.674} \\
      \bottomrule
      \end{tabular}
    \label{tab:main_rec}
\end{table*}

\section{Experiments}
\label{sec:exp}
For fair comparison, we follow common practice~\citep{mar} on ImageNet-1K~\citep{imagenet1k} at 256 $\times$ 256 resolution.

\subsection{Implementation details.} 

\textbf{\system.} The \system encoder is a ViT-B/8~\citep{vit} with registers~\citep{registers} and causal attention masks~\citep{seed,flextok}. For fair comparison, the extracted 1D tokens are 16-dimensional and normalized before being passed to the decoder following~\citep{rcg,semanticist}. The decoder is either a DiT-B/4 or DiT-L/2~\citep{dit}, which are denoted as \system-B and \system-L respectively, operating on the latent space of a frozen, publicly available KL-16 MAR-VAE~\citep{mar} to reduce computation. Both the encoder and decoder are trained from scratch on ImageNet-1K~\citep{imagenet1k} training split. Besides, we utilize DINOv2-B/16~\citep{dinov2} as the VFM of REPA and \ourrepa, and the loss weights for $\mathcal{L}_{REPA}$, and $\mathcal{L}_{REPA-A}$ are set to 1.0 and 0.8, respectively.

\begin{table*}[!ht]
      \centering
      \tablestyle{10.5pt}{1.0}
      \caption{ \textbf{Class-conditional generation results on ImageNet-1K 256 $\times$ 256 benchmark.} ``\#Param.'' denotes the parameters of generator, ``Token'' and ``Step'' indicates the number of tokens (learned tokens in tokenization training / used tokens in AR modeling) and steps used for generation, respectively. ``$\downarrow$'' or ``$\uparrow$'' denote lower or higher values are better. $\star$: trained by the official codebase~\cite{semanticist}.}
      
      \begin{tabular}{llccccc}
      \toprule
      Method & Generator &\#Param. & Token & Step  & gFID$\downarrow$ & IS$\uparrow$   \\
      \midrule
      \multicolumn{7}{c}{\textcolor{gray}{\textit{2D autoregressive models}}} \\ 
        VQGAN~\cite{vqgan} & Tam. Trans.~\cite{vqgan} & 1.4B & 256 & 256 & 15.78 & 74.3 \\
        RQ-VAE~\cite{rq-vae} & RQ-Trans.~\cite{rq-vae} & 3.8B & 256 & 68 & 7.55 & 134.0 \\
        Causal MAR~\cite{mar} & MAR-L~\cite{mar} & 481M & 256 & 256 & 4.07 & 232.4 \\
        LlamaGen~\cite{llamagen} & LlamaGen-L~\cite{llamagen} & 343M & 256 & 256 & 3.80 & 248.3 \\
        VAR~\cite{var} & VAR-d16~\cite{var} & 310M & 680 & 10 & 3.30 & 274.4 \\
      \midrule
      \multicolumn{7}{c}{\textcolor{gray}{\textit{1D masked-prediction models}}} \\ 
       
        FlowMo-Lo-256~\cite{flowmo} & MaskGiT-L~\cite{maskgit} & 227M & 256 & - & 4.30 & 274.0 \\
        TiTok-L-32~\cite{titok} & MaskGiT-L~\cite{maskgit} & 227M & 32 & 8 & 2.77 & 194.0 \\
        TiTok-S-128~\cite{titok} & MaskGiT-L~\cite{maskgit} & 227M & 128 & 64 & 1.97 & 281.8 \\
        \midrule
      \multicolumn{7}{c}{\textcolor{gray}{\textit{1D autoregressive models}}} \\ 
        FlexTok d12-d12~\cite{flextok} & AR Trans. & 1.3B & 256 / 32 & 32 & 3.83 & - \\
        SpectralAR-64~\cite{spectralar} & VAR~\cite{var} & 310M & 64 & 64 & 3.02 & 282.2 \\
        Semanticist-L-256~\cite{semanticist} & $\epsilon$LlamaGen-L~\cite{semanticist} & 343M & 256 / 32 & 32 & 2.57 & 260.9 \\
        Semanticist-L-128$^\star$~\cite{semanticist} & $\epsilon$LlamaGen-L~\cite{semanticist} & 343M & 128 / 32 & 128 & 3.33 & 251.1 \\
        Semanticist-L-128$^\star$~\cite{semanticist} & $\epsilon$LlamaGen-L~\cite{semanticist} & 343M & 128 / 128 & 128 & 4.06 & 237.2 \\
        \rowcolor{OliveGreen!5}
        \system-L-32 & $\epsilon$LlamaGen-L~\cite{semanticist} & 343M & 32  & 32 & 3.40 & 288.6 \\
        \rowcolor{OliveGreen!5}
        \system-L-64 & $\epsilon$LlamaGen-L~\cite{semanticist} & 343M & 64 & 64 & 3.01 & 280.5 \\
        \rowcolor{OliveGreen!5}
        \system-L-128 & $\epsilon$LlamaGen-L~\cite{semanticist} & 343M & 128 & 128 & 2.95 & 269.2 \\
      \bottomrule
      \end{tabular}
    \label{tab:main_gen}
\end{table*}

\textbf{Autoregressive modeling.}
Following~\citep{semanticist}, we evaluate frozen \system by training autoregressive generators LlamaGen~\citep{llamagen} with a diffusion loss~\citep{mar}. At inference, we adopt a Classifier-Free Guidance (CFG) schedule following~\citep{muse,mar,semanticist} without temperature sampling. Additional details are provided in \cref{apend:hyper}.

\subsection{Reconstruction}
We report reconstruction FID~\citep{fid} (distributional dissimilarity), PSNR (pixel-wise MSE), and SSIM~\citep{ssim} (perceptual similarity) on the ImageNet-1K validation set at 256 $\times$ 256 resolution. We evaluate five variants: \system-B with 256 1D tokens, \system-L with 32, 64, 128 and 256 tokens. The results are compared against state-of-the-art variants with comparable latent spaces and model sizes. As shown in \cref{tab:main_rec}, among diffusion autoencoders, \system-L-256 achieves superior PSNR and SSIM, with SSIM significantly outperforming the 945M FlowMo-Lo-256, while also reaching competitive rFID with less than half the training epochs compared with Semanticist-L-256. Remarkably, \system-B-256 attains comparable results with 80 epochs, demonstrating the training efficiency of \system.

Notably, \system-L-256$^\dagger$ achieves the best PSNR and SSIM among one-step 1D tokenizers, demonstrating its flexibility in sampling: it supports fast one-step sampling while also benefiting from multi-step sampling for improved reconstruction. However, although its rFID surpasses VQGAN~\citep{vqgan}—a classic 2D tokenizer—by three points, it still lags behind modern tokenizers. This gap arises because those methods rely on more challenging objectives (\eg, GAN~\citep{gan} loss) and complex training recipes~\citep{titok,oned-piece}, whereas \system is not specifically optimized for one-step sampling but attains comparable results as a byproduct.

\begin{table}[!ht]
  \centering
  \caption{\textbf{Ablation on technique designs.}}
  \begin{subtable}[t]{0.6\linewidth}
      \tablestyle{4pt}{1.0}
      \caption{ \textbf{Ablation on training recipe.} FID@$n$: n-step sampling.}
      
    \begin{tabular}{lccc}
      \toprule
      Method &  rFID@1 & rFID@25  & gFID \\
      \midrule
        $\mathcal{L}_{RF}$ in \cref{eq:loss_rf} & 183.69 & 1.81 & 19.67\\
        + $\mathcal{L}_{MF}$ in \cref{eq:loss_mf} & 4.71 & 1.90 & 24.39 \\
        + $\mathcal{L}_{REPA}$ in \cref{eq:loss_repa} & 4.31 & 1.71 & 17.92 \\
        + $\mathcal{L}_{REPA-A}$ in \cref{eq:loss_repa_a} & 3.92 & 1.15 & 13.54 \\
        + Selecting tokens in $[r, t]$  & 4.89 & 1.17 & 4.91 \\
      \bottomrule
      \end{tabular}
    \label{tab:abl_recipe}
\end{subtable}
  \hfill
  \begin{subtable}[t]{0.38\linewidth}

      \tablestyle{8pt}{1.013}
      \caption{ \textbf{Ablation on the causality and balance of 1D visual tokens.}  }
      
      \begin{tabular}{lccc}
      \toprule
      Select & Token&rFID & gFID  \\
      \midrule
        \rowcolor{OliveGreen!5}
         $[r, t]$ & 256 &1.17 & 4.91 \\
        All & 256 & 1.15 & 13.54 \\
        First $k$ & 256 & 1.37 & 9.21 \\
        \rowcolor{gray!10}
        First $k$ & 128 & 5.32 & 7.49 \\
      \bottomrule
      \end{tabular}
    \label{tab:abl_causal_balance}
\end{subtable}

\end{table}

\subsection{Autoregressive generation}
\begin{figure*}[ht!]
    \centering
    \includegraphics[width=1.0\linewidth]{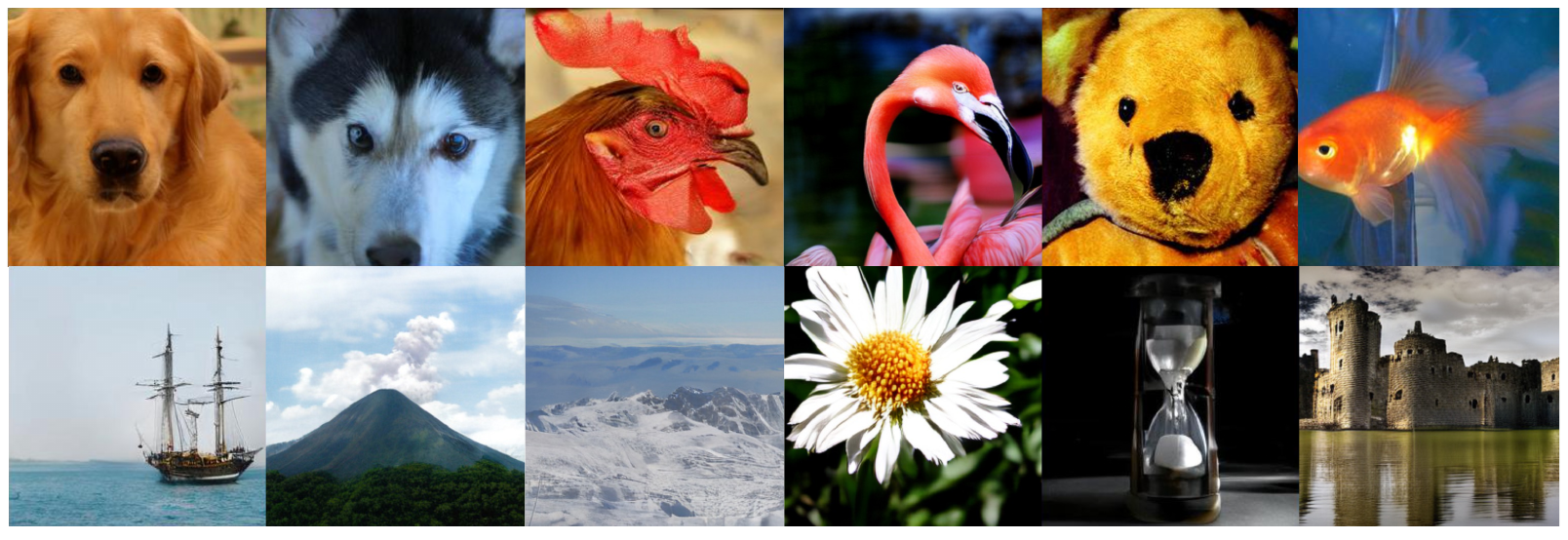}
    \caption{\textbf{Qualitative Results.}  256$\times$256 generated images on ImageNet-1K with \system-L-32.}
    \label{fig:gen}
\end{figure*}

Following common practice~\citep{mar}, we report generation FID and IS~\citep{is} (image quality and class diversity)  with evaluation suite provided by \cite{adm}. For fair comparison and efficient training, we train $\epsilon$LlamaGen-L, \ie, standard LlamaGen~\citep{llamagen} with the diffusion loss~\citep{mar} modified by \cite{semanticist}, for 400 epochs. As shown in \cref{tab:main_gen}, \system attains comparable gFID and IS scores to the state-of-the-art tokenizers, with far fewer tokenization training epochs ($160$ \vs\ $300+$), demonstrating its capability to learn 1D causal tokens well-suited for standard autoregressive modeling. We also present qualitative visualizations in \cref{fig:gen}. 

It is worth noting that training a state-of-the-art visual AR generative model is computationally expensive, and there are complicated interplays between visual tokenizers and AR generators such as the dimension of visual tokens and the accumulation of errors caused by the increasing number of tokens~\cite{llamagen,mar,magvitv2}, which are beyond the scope of this work. Instead, we focus on building a novel 1D tokenizer that captures visual causality and validating its advantages on AR modeling on ImageNet-1K~\cite{imagenet1k} under fair comparison. We leave training on larger datasets and evaluation on a broader range of tasks to the future work.

\begin{figure*}[ht!]
    \centering
    \includegraphics[width=1.0\linewidth]{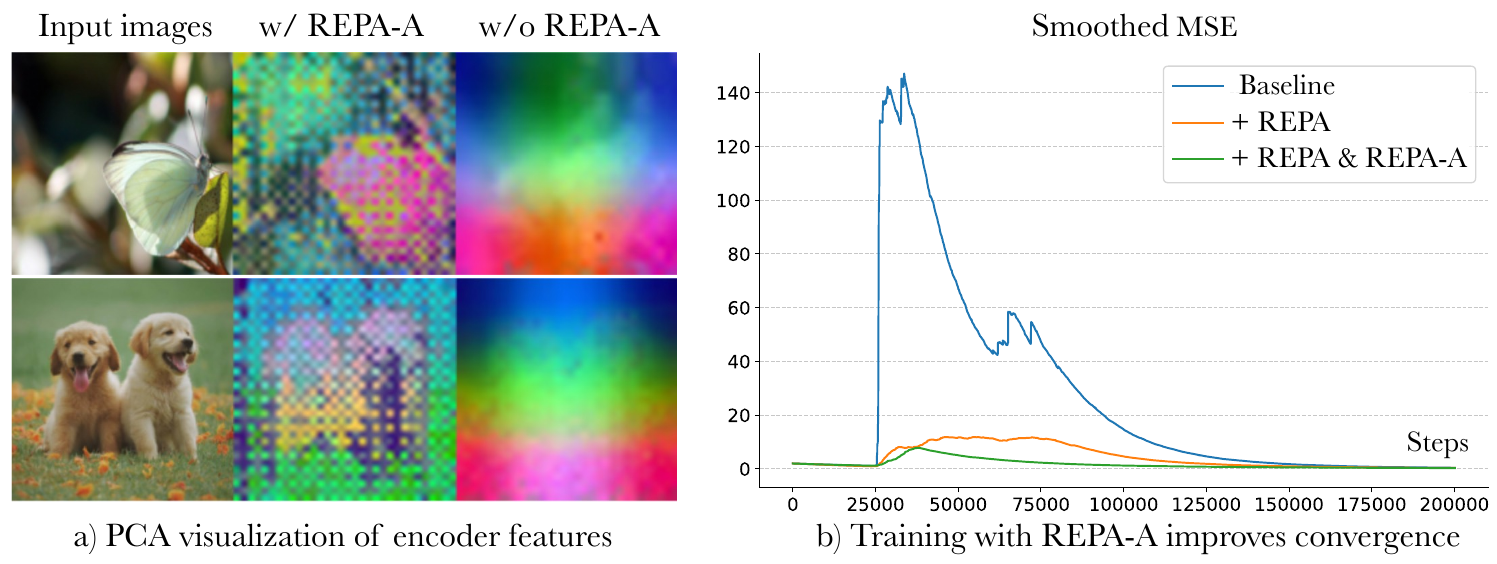}
    \caption{\textbf{Effectiveness of our \ourrepa.} a) We apply principal component analysis (PCA) to visualize image features from the \system encoder. b) Training curves of the smoothed MSE between prediction and target, with the MeanFlow loss ($\mathcal{L}_{MF}$) added at 25K steps.}
    \label{fig:visual1}
\end{figure*}

\subsection{Ablation study}
We conduct ablation studies on the smaller \system-B-256 models trained with 80 epochs.

\textbf{Improved training recipe.} We present a roadmap from the conventional diffusion autoencoder with na\"ive decoder to our \system step by step in \cref{tab:abl_recipe}. Traditional DiT decoders lack one-step sampling capability, but equipping them with the MeanFlow objective enables reasonable one-step results. Both REPA and REPA-A accelerate convergence and enhance performance. Moreover, optimizing MeanFlow objective on 1D tokens selected from a time interval $[r,t]$ allows the model to learn visual causality for AR modeling, at the cost of a slight performance drop in reconstruction.

\textbf{Causality and balance matter in AR modeling.} We evaluate three variants of 1D token selection: (1) selecting tokens within an interval $[r,t]$ (our default setting); (2) selecting all tokens; and (3) selecting the first $k$ tokens. For the third variant, we train two AR models using either all 256 tokens or only the first 128 tokens. As shown in \cref{tab:abl_causal_balance}, \system achieves the best gFID. Non-causal tokens hinder AR modeling, and, consistent with \cite{flextok, semanticist}, imbalance reduces the contribution of later tokens—an issue that \system fundamentally addresses without requiring additional re-weighting mechanism~\citep{selftok}.

\textbf{\ourrepa stabilizes training and improves performance.} As shown in \cref{fig:visual1} a), \ourrepa makes encoder features more informative and discriminative, helping the registers capture richer content. In \cref{fig:visual1} b), \ourrepa mitigates the loss spike at 25K steps when the MeanFlow loss is introduced, stabilizing decoder training and improving overall performance.

\section{Conclusion}
We presented \system, a novel 1D causal image tokenizer to bridge the gap between autoregressive language models and vision models. By binding the average velocity field in the MeanFlow objective to the corresponding 1D token segments, we enabled the diffusion autoencoder to learn visual causality along the flow path while supporting one-step sampling. Furthermore, we proposed an advanced regularization method \ourrepa, which effectively stabilized and accelerated the training of the autoencoder. Experiments demonstrated that we achieved state-of-the-art PSNR and SSIM on ImageNet reconstruction, and obtaining comparable results on the class-conditional generation.

\textbf{Acknowledgements.} This work is supported by the National Natural Science Foundation of China (Grant No. 62521004) and the New Cornerstone Science Foundation through the XPLORER PRIZE.
\section*{Appendix}
\section{More implementation details}
\newcommand{\fakeparagraphre}[1]{\noindent\textbf{#1}}

\label{apend:hyper}
\begin{table}[!ht]
    \centering
    \tablestyle{1pt}{1.2}
    \caption{Detailed configuration of \system-B and \system-L for tokenization and AR modeling.}
    \begin{tabular}{l c c c}
        \toprule
         Training Config  & \system-B & \system-L & AR modeling \\
         \hline
         Optimizer & \multicolumn3{c}{AdamW}  \\
         Peak learning rate & \multicolumn{2}{c}{$1\times10^{-4}$} &  $5\times10^{-5}$\\
         Minimum learning rate & \multicolumn{3}{c}{$0$} \\
         Learning rate schedule & \multicolumn{2}{c}{cosine decay} & constant \\
         Batch size & \multicolumn{2}{c}{1024} & 2048\\
         Weight decay & \multicolumn{3}{c}{0.05} \\
         Epochs & 80 & 160 & 400 \\
         Warmup epochs & \multicolumn{2}{c}{0} & 96 \\
         Gradient clipping & \multicolumn{3}{c}{3.0} \\
         EMA & \multicolumn{3}{c}{0.999} \\
         \bottomrule
    \end{tabular}
    \label{tab:hyper}
\end{table}
Training setup follows \cite{semanticist}, with detailed hyperparameters in \cref{tab:hyper}. For reconstruction, we disable CFG in one-step sampling, and apply CFG with a scale of 2.0 in 25-step sampling. For 80-epoch training, we introduced the MeanFlow objective at epoch 10 and the selecting mechanism at epoch 40; for 160-epoch training, these corresponded to epochs 20 and 80, respectively. For generation, we do not use CFG with \system, and the CFG of AR model is the same as MUSE~\citep{muse}, MAR~\citep{mar} and Semanticist~\citep{semanticist}, whici tunes down the guidance scale of small-indexed tokens to improve the diversity of generated sample.

\section{More experiments}

\fakeparagraphre{\system-VQ.} To evaluate the effectiveness of our approach with VQ and to avoid the cumulative errors introduced by causal MAR~\cite{mar} when the number of tokens increases, we conduct a straightforward comparison experiment with LlamaGen~\cite{llamagen}. Specifically, we integrate FSQ into CaTok without modifying any part of the original training recipe. As shown in \cref{tab:exp_vq}, because we do not perform hyperparameter tuning tailored for VQ training, nor incorporate additional techniques such as perceptual losses or post-training~\cite{llamagen,flowmo}, CaTok-VQ performs significantly worse than LlamaGen’s tokenizer in terms of rFID (3.81 \vs 2.19). However, due to the inherent causality of CaTok-VQ’s visual tokens—which is advantageous for autoregressive modeling—its AR generation performance surpasses that of LlamaGen (3.35 \vs 3.80), which further demonstrates the superior effectiveness of our approach. We believe that improved training of the VQ tokenizer, along with larger autoregressive models, can lead to further gains in generation performance.
\vspace{1.5mm}

\begin{table}[!ht]
      \tablestyle{3pt}{1.0}
      \caption{ \textbf{Reconstruction and generation results of \system-VQ}  }
      \begin{tabular}{lcccccc}
      \toprule
      Method & \#Param. & Token & Step & rFID & gFID  \\
      \midrule
         LlamaGen & 343M & 256 & 256 & 2.19 & 3.80 \\
        \rowcolor{OliveGreen!6}
         \system-LlamaGen & 343M & 256 & 256 & 3.81 & 3.35 \\
      \bottomrule
      \end{tabular}
    \label{tab:exp_vq}
\end{table}

\begin{table}[h!]
\centering
\begin{tabular}{lcccc} \toprule
 (w/o CFG) & \makecell{Learned tokens \\ in tokenization training}  & \makecell{Used tokens \\ in AR modeling} & gFID & IS \\ 
\midrule
Semanticist-L-256 & 256 & 256 & 7.60 & 121.5 \\
\rowcolor{OliveGreen!5}
\system-L-256 & 256 & 256 & \textbf{5.52} & \textbf{153.9}\\
Semanticist-L-256 & 256 & 32 & 4.96 & 147.4 \\
\rowcolor{OliveGreen!5}
\system-L-256$^{\dagger}$ & 256 & 32 & \textbf{4.77} & \textbf{165.2} \\
\bottomrule
\end{tabular}%
\end{table}

\fakeparagraphre{Apple-to-apple comparison with Semanticist without CFG.} We conduct a comparison with Semanticist under matched settings without CFG: we train an AR model using the official Semanticist tokenizer checkpoint with 256 tokens and directly evaluate the official 32-token checkpoint in the no-CFG setting. $\dagger$: we freeze the ViT encoder and fine-tune the DiT with nested dropout.
\vspace{1.5mm}

\fakeparagraphre{Extensions on the REPA encoder, latent space, and image resolution.
} \system exhibits consistent reconstruction behavior across different REPA teacher~\cite{dinov3,siglip2} and latent spaces~\cite{sd3,mar}, and the reconstruction drop mainly stems from DiT re-compressing the latent space and can be alleviated by reducing the DiT patch size to smaller. Moreover, the DiT can naturally generalize to higher resolutions via a training-free patchwise diffuse-and-blend strategy (as in FlowMo~\cite{flowmo}). Since MAR-VAE is not trained at 512×512, we replace it with SD-VAE for training at 512 resolution, and adopt the high-resolution timestep shift strategy used in SD3.
\begin{table}[h!]
\centering
\tablestyle{15pt}{1.0}

\begin{tabular}{lccc} \toprule
 \system-B-256 on ImageNet & rFID  & PSNR & SSIM  \\
 \midrule
 DINOv3 &  1.16 & 22.43 &  0.674  \\ 
 SigLIP2 &  1.12 & 21.96 &  0.657  \\ 
\hline
 MAR-VAE w/ DiT-B/2 &  0.99 & 22.69 &  0.672  \\ 
 SD-VAE &  1.34 & 21.99 &  0.658  \\ 
\hline
 \textbf{512 resolution} (training-free) &  0.60 & 27.74 & 0.778 \\ 
 \textbf{512 resolution} (trained w/ SD-VAE) &  1.07 & 24.92 & 0.705  \\ 
\bottomrule
\end{tabular}%
\end{table}

\fakeparagraphre{Additional dataset.}
To further evaluate the generalization ability of \system beyond ImageNet, we conduct additional experiments on the COCO-val-5K dataset~\cite{coco}. \system achieves consistent performance on COCO-val-5K, indicating that the learned representations generalize well to datasets with different distributions.
\begin{table}[h!]
\centering
\tablestyle{15pt}{1.0}
\begin{tabular}{lccc} \toprule
 Models on COCO-5K & rFID  & PSNR & SSIM  \\ 
 \midrule
 LlamaGen-16x16 & 8.11 & 20.42 & 0.678 \\
 Semanticist-L-256 & 5.64 & 21.36 & 0.640\\
\rowcolor{OliveGreen!5}
 \system-L-256 & \textbf{4.78} & \textbf{22.43} & \textbf{0.690}\\
\bottomrule
\end{tabular}%
\end{table}

\fakeparagraphre{Non-autoregressive generator.}
To further study the applicability of \system under different generation paradigms, we replace the autoregressive generator with a non-autoregressive generator, $\epsilon$MaskGiT, and evaluate the model under 8-step sampling. 
Under this setting, \system achieves better performance than TiTok, indicating that the proposed tokenizer is compatible with both autoregressive and mask-based generation. 
Furthermore, a variant of \system without token dropout still yields improved performance, suggesting that the introduced visual causality also benefits non-autoregressive modeling.

\begin{table}[h!]
\centering
\begin{tabular}{lcccc} \toprule
Tokenizer & Generator & Step & gFID & IS \\ 
\midrule
TiTok-L-32 & MaskGiT-L & 8 & 2.77 & 194.0 \\
\rowcolor{OliveGreen!5}
\system-L-32 w/o causality & $\epsilon$MaskGiT-L & 8 & 3.26 & 210.4
\\
\rowcolor{OliveGreen!5}
\system-L-32 & $\epsilon$MaskGiT-L & 8 & \textbf{2.69} & \textbf{223.7}
 \\
\bottomrule
\end{tabular}%
\end{table}

\clearpage

\bibliographystyle{plainnat}
\bibliography{main}

@article{meanflow,
  title={Mean flows for one-step generative modeling},
  author={Geng, Zhengyang and Deng, Mingyang and Bai, Xingjian and Kolter, J Zico and He, Kaiming},
  journal={arXiv preprint arXiv:2505.13447},
  year={2025}
}

@inproceedings{ddt,
  title={Generative Multimodal Pretraining with Discrete Diffusion Timestep Tokens},
  author={Pan, Kaihang and Lin, Wang and Yue, Zhongqi and Ao, Tenglong and Jia, Liyu and Zhao, Wei and Li, Juncheng and Tang, Siliang and Zhang, Hanwang},
  booktitle={CVPR},
  year={2025}
}

@inproceedings{flextok,
  title={FlexTok: Resampling Images into 1D Token Sequences of Flexible Length},
  author={Bachmann, Roman and Allardice, Jesse and Mizrahi, David and Fini, Enrico and Kar, O{\u{g}}uzhan Fatih and Amirloo, Elmira and El-Nouby, Alaaeldin and Zamir, Amir and Dehghan, Afshin},
  booktitle={ICML},
  year={2025}
}

@inproceedings{flowmo,
  title={Flow to the mode: Mode-seeking diffusion autoencoders for state-of-the-art image tokenization},
  author={Sargent, Kyle and Hsu, Kyle and Johnson, Justin and Fei-Fei, Li and Wu, Jiajun},
  booktitle={ICCV},
  year={2025}
}

@inproceedings{semanticist,
  title={"Principal Components" Enable A New Language of Images},
  author={Wen, Xin and Zhao, Bingchen and Elezi, Ismail and Deng, Jiankang and Qi, Xiaojuan},
  booktitle={ICCV},
  year={2025}
}

@article{selftok,
  title={Selftok: Discrete Visual Tokens of Autoregression, by Diffusion, and for Reasoning},
  author={Wang, Bohan and Yue, Zhongqi and Zhang, Fengda and Chen, Shuo and Bi, Li'an and Zhang, Junzhe and Song, Xue and Chan, Kennard Yanting and Pan, Jiachun and Wu, Weijia and others},
  journal={arXiv preprint arXiv:2505.07538},
  year={2025}
}

@inproceedings{var,
  title={Visual autoregressive modeling: Scalable image generation via next-scale prediction},
  author={Tian, Keyu and Jiang, Yi and Yuan, Zehuan and Peng, Bingyue and Wang, Liwei},
  booktitle={NeurIPS},
  year={2024}
}

@inproceedings{repa-e,
  title={Repa-e: Unlocking vae for end-to-end tuning with latent diffusion transformers},
  author={Leng, Xingjian and Singh, Jaskirat and Hou, Yunzhong and Xing, Zhenchang and Xie, Saining and Zheng, Liang},
  booktitle={ICCV},
  year={2025}
}

@inproceedings{repa,
  title={Representation alignment for generation: Training diffusion transformers is easier than you think},
  author={Yu, Sihyun and Kwak, Sangkyung and Jang, Huiwon and Jeong, Jongheon and Huang, Jonathan and Shin, Jinwoo and Xie, Saining},
  booktitle={ICLR},
  year={2024}
}

@inproceedings{dit,
  title={Scalable diffusion models with transformers},
  author={Peebles, William and Xie, Saining},
  booktitle={ICCV},
  year={2023}
}

@inproceedings{rectified-flow,
  title={Flow Straight and Fast: Learning to Generate and Transfer Data with Rectified Flow},
  author={Liu, Xingchao and Gong, Chengyue and others},
  booktitle={ICLR},
  year={2022}
}

@inproceedings{flowmatching,
  title={Flow matching for generative modeling},
  author={Lipman, Yaron and Chen, Ricky TQ and Ben-Hamu, Heli and Nickel, Maximilian and Le, Matt},
  booktitle={ICLR},
  year={2022}
}

@article{simplear,
  title={Simplear: Pushing the frontier of autoregressive visual generation through pretraining, sft, and rl},
  author={Wang, Junke and Tian, Zhi and Wang, Xun and Zhang, Xinyu and Huang, Weilin and Wu, Zuxuan and Jiang, Yu-Gang},
  journal={arXiv preprint arXiv:2504.11455},
  year={2025}
}

@inproceedings{mar,
  title={Autoregressive image generation without vector quantization},
  author={Li, Tianhong and Tian, Yonglong and Li, He and Deng, Mingyang and He, Kaiming},
  booktitle={NeurIPS},
  year={2024}
}

@article{gpt4,
  title={Gpt-4 technical report},
  author={Achiam, Josh and Adler, Steven and Agarwal, Sandhini and Ahmad, Lama and Akkaya, Ilge and Aleman, Florencia Leoni and Almeida, Diogo and Altenschmidt, Janko and Altman, Sam and Anadkat, Shyamal and others},
  journal={arXiv preprint arXiv:2303.08774},
  year={2023}
}

@misc{gpt5,
  author={OpenAI},
  title={GPT-5 System Card},
  howpublished={\url{https://openai.com/index/gpt-5-system-card/}},
  year={2025}
}

@misc{sora,
  author={OpenAI},
  title={Sora System Card},
  howpublished={\url{https://openai.com/index/sora-system-card/}},
  year={2024}
}

@article{llama3,
  title={The llama 3 herd of models},
  author={Grattafiori, Aaron and Dubey, Abhimanyu and Jauhri, Abhinav and Pandey, Abhinav and Kadian, Abhishek and Al-Dahle, Ahmad and Letman, Aiesha and Mathur, Akhil and Schelten, Alan and Vaughan, Alex and others},
  journal={arXiv preprint arXiv:2407.21783},
  year={2024}
}

@article{qwen3,
  title={Qwen3 technical report},
  author={Yang, An and Li, Anfeng and Yang, Baosong and Zhang, Beichen and Hui, Binyuan and Zheng, Bo and Yu, Bowen and Gao, Chang and Huang, Chengen and Lv, Chenxu and others},
  journal={arXiv preprint arXiv:2505.09388},
  year={2025}
}

@article{deepseek,
  title={Deepseek-v3 technical report},
  author={Liu, Aixin and Feng, Bei and Xue, Bing and Wang, Bingxuan and Wu, Bochao and Lu, Chengda and Zhao, Chenggang and Deng, Chengqi and Zhang, Chenyu and Ruan, Chong and others},
  journal={arXiv preprint arXiv:2412.19437},
  year={2024}
}

@article{gemini,
  title={Gemini 2.5: Pushing the frontier with advanced reasoning, multimodality, long context, and next generation agentic capabilities},
  author={Comanici, Gheorghe and Bieber, Eric and Schaekermann, Mike and Pasupat, Ice and Sachdeva, Noveen and Dhillon, Inderjit and Blistein, Marcel and Ram, Ori and Zhang, Dan and Rosen, Evan and others},
  journal={arXiv preprint arXiv:2507.06261},
  year={2025}
}

@inproceedings{lvm,
  title={Sequential modeling enables scalable learning for large vision models},
  author={Bai, Yutong and Geng, Xinyang and Mangalam, Karttikeya and Bar, Amir and Yuille, Alan L and Darrell, Trevor and Malik, Jitendra and Efros, Alexei A},
  booktitle={CVPR},
  year={2024}
}

@inproceedings{vqgan,
  title={Taming transformers for high-resolution image synthesis},
  author={Esser, Patrick and Rombach, Robin and Ommer, Bjorn},
  booktitle={CVPR},
  year={2021}
}

@inproceedings{dalle,
  title={Zero-shot text-to-image generation},
  author={Ramesh, Aditya and Pavlov, Mikhail and Goh, Gabriel and Gray, Scott and Voss, Chelsea and Radford, Alec and Chen, Mark and Sutskever, Ilya},
  booktitle={ICML},
  year={2021},
}

@inproceedings{vqvae-2,
  title={Generating diverse high-fidelity images with vq-vae-2},
  author={Razavi, Ali and Van den Oord, Aaron and Vinyals, Oriol},
  booktitle={NeurIPS},
  year={2019}
}

@article{llamagen,
  title={Autoregressive model beats diffusion: Llama for scalable image generation},
  author={Sun, Peize and Jiang, Yi and Chen, Shoufa and Zhang, Shilong and Peng, Bingyue and Luo, Ping and Yuan, Zehuan},
  journal={arXiv preprint arXiv:2406.06525},
  year={2024}
}

@inproceedings{magvitv2,
  title={Language Model Beats Diffusion--Tokenizer is Key to Visual Generation},
  author={Yu, Lijun and Lezama, Jos{\'e} and Gundavarapu, Nitesh B and Versari, Luca and Sohn, Kihyuk and Minnen, David and Cheng, Yong and Birodkar, Vighnesh and Gupta, Agrim and Gu, Xiuye and others},
  booktitle={ICLR},
  year={2024}
}

@article{emu3,
  title={Emu3: Next-token prediction is all you need},
  author={Wang, Xinlong and Zhang, Xiaosong and Luo, Zhengxiong and Sun, Quan and Cui, Yufeng and Wang, Jinsheng and Zhang, Fan and Wang, Yueze and Li, Zhen and Yu, Qiying and others},
  journal={arXiv preprint arXiv:2409.18869},
  year={2024}
}

@inproceedings{infinity,
  title={Infinity: Scaling bitwise autoregressive modeling for high-resolution image synthesis},
  author={Han, Jian and Liu, Jinlai and Jiang, Yi and Yan, Bin and Zhang, Yuqi and Yuan, Zehuan and Peng, Bingyue and Liu, Xiaobing},
  booktitle={CVPR},
  year={2025}
}

@inproceedings{ddpm,
  title={Denoising diffusion probabilistic models},
  author={Ho, Jonathan and Jain, Ajay and Abbeel, Pieter},
  booktitle={NeurIPS},
  year={2020}
}

@inproceedings{score-based,
  title={Score-based generative modeling through stochastic differential equations},
  author={Song, Yang and Sohl-Dickstein, Jascha and Kingma, Diederik P and Kumar, Abhishek and Ermon, Stefano and Poole, Ben},
  booktitle={ICLR},
  year={2021}
}

@article{qwen-image,
  title={Qwen-image technical report},
  author={Wu, Chenfei and Li, Jiahao and Zhou, Jingren and Lin, Junyang and Gao, Kaiyuan and Yan, Kun and Yin, Sheng-ming and Bai, Shuai and Xu, Xiao and Chen, Yilei and others},
  journal={arXiv preprint arXiv:2508.02324},
  year={2025}
}

@article{rar,
  title={Randomized autoregressive visual generation},
  author={Yu, Qihang and He, Ju and Deng, Xueqing and Shen, Xiaohui and Chen, Liang-Chieh},
  journal={arXiv preprint arXiv:2411.00776},
  year={2024}
}

@inproceedings{nestdrop,
  title={Learning ordered representations with nested dropout},
  author={Rippel, Oren and Gelbart, Michael and Adams, Ryan},
  booktitle={ICML},
  year={2014},
}

@inproceedings{titok,
  title={An image is worth 32 tokens for reconstruction and generation},
  author={Yu, Qihang and Weber, Mark and Deng, Xueqing and Shen, Xiaohui and Cremers, Daniel and Chen, Liang-Chieh},
  booktitle={NeurIPS},
  year={2024}
}

@inproceedings{seed,
  title={Chatlaw: A multi-agent collaborative legal assistant with knowledge graph enhanced mixture-of-experts large language model},
  author={Cui, Jiaxi and Ning, Munan and Li, Zongjian and Chen, Bohua and Yan, Yang and Li, Hao and Ling, Bin and Tian, Yonghong and Yuan, Li},
  booktitle={ICLR},
  year={2024}
}

@inproceedings{diff-ae,
  title={Diffusion autoencoders: Toward a meaningful and decodable representation},
  author={Preechakul, Konpat and Chatthee, Nattanat and Wizadwongsa, Suttisak and Suwajanakorn, Supasorn},
  booktitle={CVPR},
  year={2022}
}

@inproceedings{cdiff-ae,
  title={Lossy image compression with conditional diffusion models},
  author={Yang, Ruihan and Mandt, Stephan},
  booktitle={NeurIPS},
  year={2023}
}

@inproceedings{registers,
  title={Vision Transformers Need Registers},
  author={Timoth{\'e}e Darcet and Maxime Oquab and Julien Mairal and Piotr Bojanowski},
  booktitle={ICLR},
  year={2024},
}

@article{facm,
  title={Flow-anchored consistency models},
  author={Peng, Yansong and Zhu, Kai and Liu, Yu and Wu, Pingyu and Li, Hebei and Sun, Xiaoyan and Wu, Feng},
  journal={arXiv preprint arXiv:2507.03738},
  year={2025},
}

@inproceedings{imagenet1k,
  author={Deng, Jia and Dong, Wei and Socher, Richard and Li, Li-Jia and Kai Li and Li Fei-Fei},
  booktitle={CVPR}, 
  title={ImageNet: A large-scale hierarchical image database}, 
  year={2009}
}

@inproceedings{vit,
  title={An image is worth 16x16 words: Transformers for image recognition at scale},
  author={Dosovitskiy, Alexey and Beyer, Lucas and Kolesnikov, Alexander and Weissenborn, Dirk and Zhai, Xiaohua and Unterthiner, Thomas and Dehghani, Mostafa and Minderer, Matthias and Heigold, Georg and Gelly, Sylvain and others},
  booktitle={ICLR},
  year={2021}
}

@inproceedings{vae,
  title={Auto-encoding variational bayes},
  author={Kingma, Diederik P and Welling, Max},
  booktitle={ICLR},
  year={2014}
}

@inproceedings{va-vae,
  title={Reconstruction vs. generation: Taming optimization dilemma in latent diffusion models},
  author={Yao, Jingfeng and Yang, Bin and Wang, Xinggang},
  booktitle={CVPR},
  year={2025}
}

@inproceedings{vq-vae,
  title={Neural discrete representation learning},
  author={Van Den Oord, Aaron and Vinyals, Oriol and others},
  booktitle={NeurIPS},
  year={2017}
}

@inproceedings{rq-vae,
  title={Autoregressive image generation using residual quantization},
  author={Lee, Doyup and Kim, Chiheon and Kim, Saehoon and Cho, Minsu and Han, Wook-Shin},
  booktitle={CVPR},
  year={2022}
}

@article{openmagvit2,
  title={Open-magvit2: An open-source project toward democratizing auto-regressive visual generation},
  author={Luo, Zhuoyan and Shi, Fengyuan and Ge, Yixiao and Yang, Yujiu and Wang, Limin and Shan, Ying},
  journal={arXiv preprint arXiv:2409.04410},
  year={2024}
}

@article{maskbit,
  title={Maskbit: Embedding-free image generation via bit tokens},
  author={Weber, Mark and Yu, Lijun and Yu, Qihang and Deng, Xueqing and Shen, Xiaohui and Cremers, Daniel and Chen, Liang-Chieh},
  journal={TMLR},
  year={2024}
}

@inproceedings{maskgit,
  title={Maskgit: Masked generative image transformer},
  author={Chang, Huiwen and Zhang, Han and Jiang, Lu and Liu, Ce and Freeman, William T},
  booktitle={CVPR},
  year={2022}
}

@inproceedings{blip-2,
  title={Blip-2: Bootstrapping language-image pre-training with frozen image encoders and large language models},
  author={Li, Junnan and Li, Dongxu and Savarese, Silvio and Hoi, Steven},
  booktitle={ICML},
  year={2023},
}

@inproceedings{textok,
  title={Language-guided image tokenization for generation},
  author={Zha, Kaiwen and Yu, Lijun and Fathi, Alireza and Ross, David A and Schmid, Cordelia and Katabi, Dina and Gu, Xiuye},
  booktitle={CVPR},
  year={2025}
}

@article{ta-titok,
  title={Democratizing text-to-image masked generative models with compact text-aware one-dimensional tokens},
  author={Kim, Dongwon and He, Ju and Yu, Qihang and Yang, Chenglin and Shen, Xiaohui and Kwak, Suha and Chen, Liang-Chieh},
  journal={arXiv preprint arXiv:2501.07730},
  year={2025}
}

@article{oned-piece,
  title={One-d-piece: Image tokenizer meets quality-controllable compression},
  author={Miwa, Keita and Sasaki, Kento and Arai, Hidehisa and Takahashi, Tsubasa and Yamaguchi, Yu},
  journal={arXiv preprint arXiv:2501.10064},
  year={2025}
}

@inproceedings{alit,
  title={Adaptive length image tokenization via recurrent allocation},
  author={Duggal, Shivam and Isola, Phillip and Torralba, Antonio and Freeman, William T},
  booktitle={ICLR},
  year={2025}
}

@inproceedings{spectralar,
  title={SpectralAR: Spectral Autoregressive Visual Generation},
  author={Huang, Yuanhui and Chen, Weiliang and Zheng, Wenzhao and Duan, Yueqi and Zhou, Jie and Lu, Jiwen},
  booktitle={ICCV},
  year={2025}
}

@article{dito,
  title={Diffusion autoencoders are scalable image tokenizers},
  author={Chen, Yinbo and Girdhar, Rohit and Wang, Xiaolong and Rambhatla, Sai Saketh and Misra, Ishan},
  journal={arXiv preprint arXiv:2501.18593},
  year={2025}
}

@inproceedings{rcg,
  title={Return of unconditional generation: A self-supervised representation generation method},
  author={Li, Tianhong and Katabi, Dina and He, Kaiming},
  booktitle={NeurIPS},
  year={2024}
}

@article{dinov2,
  title={DINOv2: Learning Robust Visual Features without Supervision},
  author={Oquab, Maxime and Darcet, Timoth{\'e}e and Moutakanni, Th{\'e}o and Vo, Huy and Szafraniec, Marc and Khalidov, Vasil and Fernandez, Pierre and Haziza, Daniel and Massa, Francisco and El-Nouby, Alaaeldin and others},
  journal={TMLR},
  year={2024}
}

@inproceedings{muse,
  title={Muse: Text-to-image generation via masked generative transformers},
  author={Chang, Huiwen and Zhang, Han and Barber, Jarred and Maschinot, AJ and Lezama, Jose and Jiang, Lu and Yang, Ming-Hsuan and Murphy, Kevin and Freeman, William T and Rubinstein, Michael and others},
  booktitle={ICML},
  year={2023}
}

@inproceedings{vilex,
  title={Visual lexicon: Rich image features in language space},
  author={Wang, XuDong and Zhou, Xingyi and Fathi, Alireza and Darrell, Trevor and Schmid, Cordelia},
  booktitle={CVPR},
  year={2025}
}

@inproceedings{fid,
  title={Gans trained by a two time-scale update rule converge to a local nash equilibrium},
  author={Heusel, Martin and Ramsauer, Hubert and Unterthiner, Thomas and Nessler, Bernhard and Hochreiter, Sepp},
  booktitle={NeurIPS},
  year={2017}
}

@article{ssim,
  author={Zhou Wang and Bovik, A.C. and Sheikh, H.R. and Simoncelli, E.P.},
  journal={TIP}, 
  title={Image quality assessment: from error visibility to structural similarity}, 
  year={2004},
}

@inproceedings{is,
  title={Improved techniques for training gans},
  author={Salimans, Tim and Goodfellow, Ian and Zaremba, Wojciech and Cheung, Vicki and Radford, Alec and Chen, Xi},
  booktitle={NeurIPS},
  year={2016}
}

@inproceedings{gan,
  title={Generative adversarial networks},
  author={Goodfellow, Ian and Pouget-Abadie, Jean and Mirza, Mehdi and Xu, Bing and Warde-Farley, David and Ozair, Sherjil and Courville, Aaron and Bengio, Yoshua},
  booktitle={NeurIPS},
  year={2014},
}

@inproceedings{adm,
  title={Diffusion models beat gans on image synthesis},
  author={Dhariwal, Prafulla and Nichol, Alexander},
  booktitle={NeurIPS},
  year={2021}
}

@inproceedings{dualtoken,
  title={DualToken: Towards Unifying Visual Understanding and Generation with Dual Visual Vocabularies},
  author={Song, Wei and Wang, Yuran and Song, Zijia and Li, Yadong and Zhou, Zenan and Chen, Long and Xu, Jianhua and Wang, Jiaqi and Yu, Kaicheng},
  booktitle={ICLR},
  year={2026}
}

@inproceedings{omnitokenizer,
  title={Omnitokenizer: A joint image-video tokenizer for visual generation},
  author={Wang, Junke and Jiang, Yi and Yuan, Zehuan and Peng, Bingyue and Wu, Zuxuan and Jiang, Yu-Gang},
  booktitle={NeurIPS},
  year={2024}
}

@article{ssdd,
  title={Ssdd: Single-step diffusion decoder for efficient image tokenization},
  author={Vallaeys, Th{\'e}ophane and Verbeek, Jakob and Cord, Matthieu},
  journal={arXiv preprint arXiv:2510.04961},
  year={2025}
}

@article{dinov3,
  title={Dinov3},
  author={Sim{\'e}oni, Oriane and Vo, Huy V and Seitzer, Maximilian and Baldassarre, Federico and Oquab, Maxime and Jose, Cijo and Khalidov, Vasil and Szafraniec, Marc and Yi, Seungeun and Ramamonjisoa, Micha{\"e}l and others},
  journal={arXiv preprint arXiv:2508.10104},
  year={2025}
}

@article{siglip2,
  title={Siglip 2: Multilingual vision-language encoders with improved semantic understanding, localization, and dense features},
  author={Tschannen, Michael and Gritsenko, Alexey and Wang, Xiao and Naeem, Muhammad Ferjad and Alabdulmohsin, Ibrahim and Parthasarathy, Nikhil and Evans, Talfan and Beyer, Lucas and Xia, Ye and Mustafa, Basil and others},
  journal={arXiv preprint arXiv:2502.14786},
  year={2025}
}

@inproceedings{sd3,
  title={Scaling rectified flow transformers for high-resolution image synthesis},
  author={Esser, Patrick and Kulal, Sumith and Blattmann, Andreas and Entezari, Rahim and M{\"u}ller, Jonas and Saini, Harry and Levi, Yam and Lorenz, Dominik and Sauer, Axel and Boesel, Frederic and others},
  booktitle={ICML},
  year={2024}
}

@inproceedings{coco,
  title={Microsoft coco: Common objects in context},
  author={Lin, Tsung-Yi and Maire, Michael and Belongie, Serge and Hays, James and Perona, Pietro and Ramanan, Deva and Doll{\'a}r, Piotr and Zitnick, C Lawrence},
  booktitle={ECCV},
  year={2014},
}






\end{document}